\documentclass{nature}

\usepackage{bm}

\newcommand*{\defeq}

\usepackage{setspace}
\usepackage{amsfonts}
\usepackage{amssymb}
\usepackage{amsmath}

\usepackage{float}
\usepackage{placeins}
\usepackage{graphicx}

\usepackage[export]{adjustbox}

\usepackage[normal,normal,bf,labelformat=empty]{caption}

\newcounter{mybodyfigure}
\newcounter{myedfigure}

\newcommand{\beginbodyfigures}{\renewcommand{\thefigure}{{\themybodyfigure}}}

\newcommand{\beginedfigures}{\renewcommand{\thefigure}{{\themyedfigure}}}

\newcommand{\stepbodyfigure}{\refstepcounter{mybodyfigure}}

\DeclareRobustCommand{\bodyfigure}[1]{\stepbodyfigure\label{#1}{\themybodyfigure}}

\newcommand{\bodyfigurelabel}[1]{\bf{Figure \bodyfigure{#1}:}}

\makeatletter
\g@addto@macro\caption@prepareslc{%
  \renewcommand{\stepbodyfigure}{\caption@l@stepcounter{mybodyfigure}}}
\makeatother

\newcommand{\stepedfigure}{\refstepcounter{myedfigure}}

\makeatletter
\g@addto@macro\caption@prepareslc{%
  \renewcommand{\stepedfigure}{\caption@l@stepcounter{myedfigure}}}
\makeatother

\usepackage[super,comma,sort&compress]{natbib}

\usepackage{hyperref}
\usepackage{verbatim}

\usepackage{ulem} %provides various types of underlining that can stretch between words and be broken across lines
\usepackage{rotating} % rotate any object by arbitrary angle
\usepackage{soul}%h y p h e n -a t a b l e l e t t e r s p a c i n g ( s p a c i n g o u t ) , underlining and some derivatives such as overstriking and highlighting

\usepackage{color} %change color of text

\usepackage[colorinlistoftodos]{todonotes}

\makeatletter
\newsavebox\myboxA
\newsavebox\myboxB
\newlength\mylenA

\newcommand*\xoverline[2][0.75]{%
    \sbox{\myboxA}{$\m@th#2$}%
    \setbox\myboxB\null% Phantom box
    \ht\myboxB=\ht\myboxA%
    \dp\myboxB=\dp\myboxA%
    \wd\myboxB=#1\wd\myboxA% Scale phantom
    \sbox\myboxB{$\m@th\overline{\copy\myboxB}$}%  Overlined phantom
    \setlength\mylenA{\the\wd\myboxA}%   calc width diff
    \addtolength\mylenA{-\the\wd\myboxB}%
    \ifdim\wd\myboxB<\wd\myboxA%
       \rlap{\hskip 0.5\mylenA\usebox\myboxB}{\usebox\myboxA}%
    \else
        \hskip -0.5\mylenA\rlap{\usebox\myboxA}{\hskip 0.5\mylenA\usebox\myboxB}%
    \fi}
\makeatother

\usepackage{booktabs}

\usepackage{float}
\usepackage{siunitx}
\usepackage{algorithm}
\usepackage{algpseudocode}  % 
\usepackage{caption}
\usepackage{mathtools, nccmath, textcomp} 
\usepackage{amsthm}
\usepackage{subfigure}

\title{MonitorVLM-v2: A Deployed Vision-Language Framework for Real-Time Safety Violation Detection}

\author{Jiang Wu$^{1}$, Sichao Wu$^{1}$, Yinsong~Ma$^{2}$, Lifang~Zheng$^{1}$, Jingliang Duan$^{1*}$}

\begin{document}

\maketitle

\begin{affiliations}
 \item School of Mechanical Engineering, University of Science and Technology Beijing
  \item The Laboratory for Computational Sensing and Robotics, Johns Hopkins University
% \author[2]{Yinsong~Ma}
% \emailauthor{yma71@jhu.edu}{Yinsong~Ma}
% \credit{Visualization, Software}
% \affiliation[2]{organization={The Laboratory for Computational Sensing and Robotics, Johns Hopkins University},
%     postcode={21218}, city={Baltimore, MD}, country={USA}}
 \item[] $^{*}$Correspondence: duanjl@ustb.edu.cn
 
\end{affiliations}

\begin{abstract}

Large vision--language models (VLMs) can reason step by step about complex visual scenes, but this open-ended, autoregressive chain-of-thought (CoT) approach is poorly suited to safety-critical, rule-governed settings such as industrial surveillance, where decisions must be bounded, deterministic, and low-latency. Because CoT inference cost scales jointly with reasoning length and the number of concurrent streams, it creates a throughput bottleneck that precludes the real-time, multistream monitoring required for industrial accountability. Here we present MonitorVLM-v2, a deployment-oriented framework that recasts VLM-based safety assessment as probabilistic inference over a finite regulatory decision space, compressing multimodal reasoning into single-step rule-ID predictions and reducing decoding from a variable-length sequence to a single token. We introduce symbolic policy optimization (SymPO), a novel contrastive policy optimization algorithm that sharpens decision boundaries within this finite symbolic space, together with an entropy-driven triage mechanism that routes uncertain predictions to human reviewers for expert confirmation. In a four-month prospective deployment across 10 concurrent camera feeds in an operational underground mining facility, MonitorVLM-v2 achieved a 19.45-fold increase in inference speed and identified 2.78 times as many confirmed violations as the site's routine manual inspection workflow, demonstrating the practical value of compressed symbolic decision-making for real-time, auditable industrial monitoring.
\end{abstract}

\beginbodyfigures
%\setcounter{mypostfigure}{0}
% \section*{Introduction}
\section{Introduction}
Large vision--language models (VLMs) have rapidly advanced multimodal intelligence from object-level perception to language-mediated reasoning. These models enable systems to describe complex scenes, infer implicit relationships, and generate explanatory conclusions from visual evidence\cite{schulze2025visual,gachter2025people,dyreborg2022safety,shah2024artificial}. This advancement has driven increasing interest in the application of VLMs to industrial monitoring and operational decision-making. However, real-world safety surveillance imposes requirements that differ fundamentally from those of open-ended visual question answering. In industrial settings, continuous multistream video must be converted into timely, reliable, and auditable rule-level safety decisions\cite{sujit2025real,zhou2024larger,alampara2025probing,jiang2025safechain} that support immediate human intervention (Fig.~\ref{fig.monitor}a).

As shown in Fig.~\ref{fig.monitor}b, existing surveillance pipelines are not well aligned with this deployment objective.
 Object detectors and action recognizers
can localize workers, equipment, protective gear and predefined motions\cite{liu2024helmet,
wang2026method,
song2024improved}, but the presence of these entities
does not by itself establish whether a procedural rule has been violated.
Image--text models such as CLIP align
visual and linguistic representations\cite{tsai2025construction,
gil2024zero,
rasheed2023fine, radford2021learning}, yet similarity scores do not provide a direct decision interface over
a structured set of regulatory hypotheses.
As a result, existing methods largely capture fragmented scene evidence, such as objects, actions, and semantic correspondences, without reliably translating this evidence into the bounded safety judgements required for operational deployment.

VLMs with chain-of-thought (CoT) prompting can, in principle, reason through complex safety scenarios step by step\cite{zheng2025learning,zhang2024multimodal,yao2025efficient,wei2022chain}. However, as illustrated in Fig.~\ref{fig.monitor}c, their autoregressive decoding is variable in length, and inference latency scales jointly with reasoning trace length and the number of concurrent streams\cite{hagendorff2023human,wu2025monitorvlm,ding2023parameter,zhang2024multimodal}. Under the coupled demands of multistream processing and per-segment regulatory evaluation, this creates a throughput bottleneck that precludes real-time deployment. Moreover, free-form textual outputs provide a weaker interface for deterministic control, auditability, and stable multistream throughput than bounded rule-level decisions. Industrial safety monitoring is therefore better framed as probabilistic inference over a finite regulatory action space: the central objective is not to produce elaborate reasoning traces, but to compress multimodal evidence into rapid, calibrated, and auditable symbolic decisions.

To address these challenges, we present MonitorVLM-v2, the first VLM framework to achieve longitudinally validated, continuously operational deployment for rule-governed industrial safety monitoring. The framework introduces three integrated innovations. First, a reasoning-to-decision compression paradigm reformulates industrial surveillance as bounded symbolic decision-making over a finite regulatory hypothesis space, mapping temporal visual segments directly to single-step rule-ID predictions and thereby decoupling inference complexity from output length. Second, we propose symbolic policy optimization (SymPO), a novel contrastive policy optimization algorithm specifically engineered for finite symbolic decision spaces, which sharpens decision boundaries by explicitly penalizing competing regulatory hypotheses under visually ambiguous conditions. Third, an entropy-guided triage mechanism routes uncertain predictions, quantified via Shannon entropy, to compact Top-3 candidate sets for expert review, establishing a scalable interface between autonomous symbolic inference and human oversight. We validated the framework through controlled experiments and a four-month prospective field deployment in a mining environment characterized by degraded visual conditions, strict safety regulations, and 10 concurrent surveillance feeds, where MonitorVLM-v2 achieved 24-hour monitoring coverage and identified 2.78 times as many confirmed violations as the site's routine manual inspection workflow.
\clearpage
\begin{figure}[!h] 
    \centering
    \includegraphics[width=0.8\textwidth]{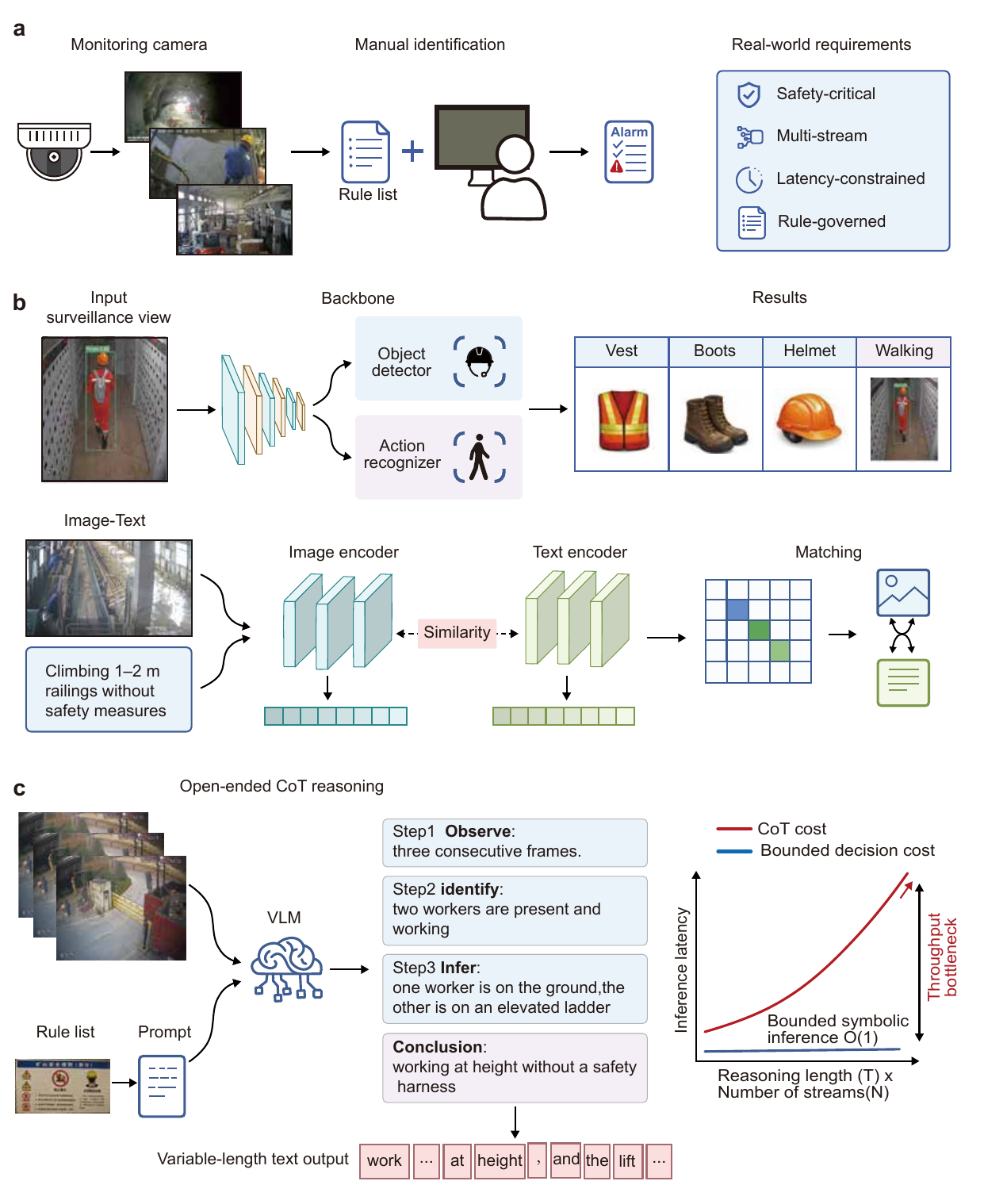}
\end{figure}

\begin{center}
    \captionsetup{type=figure, font=footnotesize}   % ← 加 font=small
    \caption{\bodyfigurelabel{fig.monitor}
    \textbf{From surveillance perception to deployment-mismatched multimodal reasoning.}
\textbf{a}, \textbf{Industrial surveillance requires rule-level safety decisions.}Continuous multistream monitoring relies on human inspectors to evaluate surveillance footage against predefined safety regulations and issue timely warnings under safety-critical, latency-constrained, multistream, and rule-governed conditions.
\textbf{b}, \textbf{Conventional hazard-recognition pipelines provide fragmented evidence.}
Object detectors and action recognizers identify entities and motions but cannot 
determine whether a procedural rule has been violated. Vision-language matching 
frameworks such as CLIP estimate image-text semantic similarity but lack a structured 
decision interface over regulatory hypotheses.
\textbf{c}, 
Open-ended VLM reasoning is expressive but deployment-mismatched. VLMs 
with CoT prompting generate step-by-step textual rationales, but autoregressive 
decoding produces variable-length outputs whose inference cost scales with both 
reasoning length and the number of concurrent streams, creating a throughput bottleneck 
(right, red curve). Bounded symbolic decision-making maps visual evidence directly to 
a finite set of rule hypotheses, achieving constant-cost $O(1)$ decoding regardless of 
stream count (right, blue line), and provides a more predictable interface for 
low-latency, auditable industrial deployment.}
\end{center}

\noindent\textbf{Reasoning-to-decision compression in bounded symbolic spaces.}

We reformulate safety monitoring as probabilistic inference over a bounded regulatory space (Fig.~\ref{fig.ACD1}a). We let \(\mathcal{R}=\{r_1,\ldots,r_K\}\) denote a finite set of operational safety regulations and \(\mathcal{Y}=\{y_0,y_1,\ldots,y_K\}\) denote the corresponding symbolic decision space, where \(y_0\) represents the nonviolation class and each \(y_k\) is associated with regulation \(r_k\). Given a temporal visual segment \(I=\{I_1,I_2,I_3\}\) and the regulatory context \(\mathcal{R}\), MonitorVLM-v2 replaces autoregressive CoT generation with a symbolic decision distribution \(\pi_\theta(y\mid I,\mathcal{R})\) over \(\mathcal{Y}\). The predicted regulatory outcome is obtained by single-step selection as follows:
\[
\hat{y}=\arg\max_{y\in\mathcal{Y}} \pi_\theta(y\mid I,\mathcal{R}).
\]
By constraining the output to a finite rule-ID space, the framework decouples online decoding from the length of natural-language reasoning traces, thereby reducing the decoding cost from \(O(T)\) to \(O(1)\), where \(T\) denotes the number of generated language tokens in an autoregressive CoT baseline.

To enable dynamic adaptation to regulatory configurations, we introduce \text{rule--token shuffling} (RTS). During training, RTS periodically permutes the rule-to-token mapping $\phi:\mathcal{R}\to\mathcal{Y}$, compelling the model to perform conditional inference over the current rule-semantic--symbol assignment rather than memorising fixed token identities. This formulation endows MonitorVLM-v2 with the capacity to generalise across dynamically reconfigured regulatory vocabularies: at each training step, the model must ground its prediction in the correspondence between visual evidence and the regulatory semantics specified by the active mapping, not in static token-index associations. During deployment, $\phi$ is fixed to a deterministic dictionary to ensure unambiguous interpretation of rule-ID outputs by downstream monitoring systems.

\noindent\textbf{Finite symbolic space optimization via contrastive policy learning.}
With reasoning compressed into a finite rule space, the central optimization challenge shifts to shaping the probability distribution over competing hypotheses (Fig.~\ref{fig.ACD1}b). Standard supervised fine-tuning (SFT) maximizes the likelihood of the ground-truth rule-ID token, but it does not explicitly suppress incorrect alternatives, thereby often producing weakly separated symbolic distributions that remain vulnerable under visually ambiguous scenes. 

To improve rule-level discrimination, we introduce symbolic policy optimization (SymPO), a contrastive algorithm specifically designed for bounded rule-ID prediction. Given the policy $\pi_\theta(y\mid I,\mathcal{R})$ and the ground-truth rule-ID $y^*$, we define the objective as follows:
\begin{equation}
\mathcal{L}_{\mathrm{SymPO}}(\theta)=
-\mathbb{E}_{(I,\mathcal{R})\sim\mathcal{D},\,y\sim\pi_\theta(\cdot|I,\mathcal{R})}
\left[R(y,y^*)\log\pi_\theta(y\mid I,\mathcal{R})\right],
\end{equation}
where $R(y,y^*)$ serves as an absolute contrastive reward and assigns $+1$ to the ground-truth $y^*$ and $-1$ to all incorrect alternatives. By assigning a negative reward to incorrectly sampled alternatives, SymPO encourages the probability mass to move away from high-confidence distractors and improves rule-level discrimination under visually ambiguous conditions.

\noindent\textbf{Uncertainty-guided human--AI collaboration.}
In complex operating environments, predictive uncertainty is an inherent property of model inference rather than a nuisance to be hidden behind a single point estimate. Because MonitorVLM-v2 outputs a probability distribution over the symbolic decision space, we quantify predictive uncertainty using Shannon entropy\cite{lin1991divergence}:
\begin{equation}
H(I,\mathcal{R})=-\sum_{y\in\mathcal{Y}}\pi_\theta(y\mid I,\mathcal{R})\log\pi_\theta(y\mid I,\mathcal{R}).
\end{equation}

We operationalize this entropy as a deployment routing signal to structure human oversight (Fig.~\ref{fig.ACD1}c). The system uses an adaptive triage strategy: low-entropy predictions are used for rapid Top-1 human confirmation, whereas high-entropy cases that indicate perceptual conflict are elevated to a compact Top-3 candidate set for expert inspection. By treating uncertainty as a measurable system signal, the framework establishes an explicit interface between autonomous symbolic inference and human supervisory judgement, thus maintaining 100\% human auditability and supporting high-throughput industrial deployment.

\clearpage
\begin{figure}[!tpbh] 
    \centering
    \includegraphics[width=0.72\textwidth]{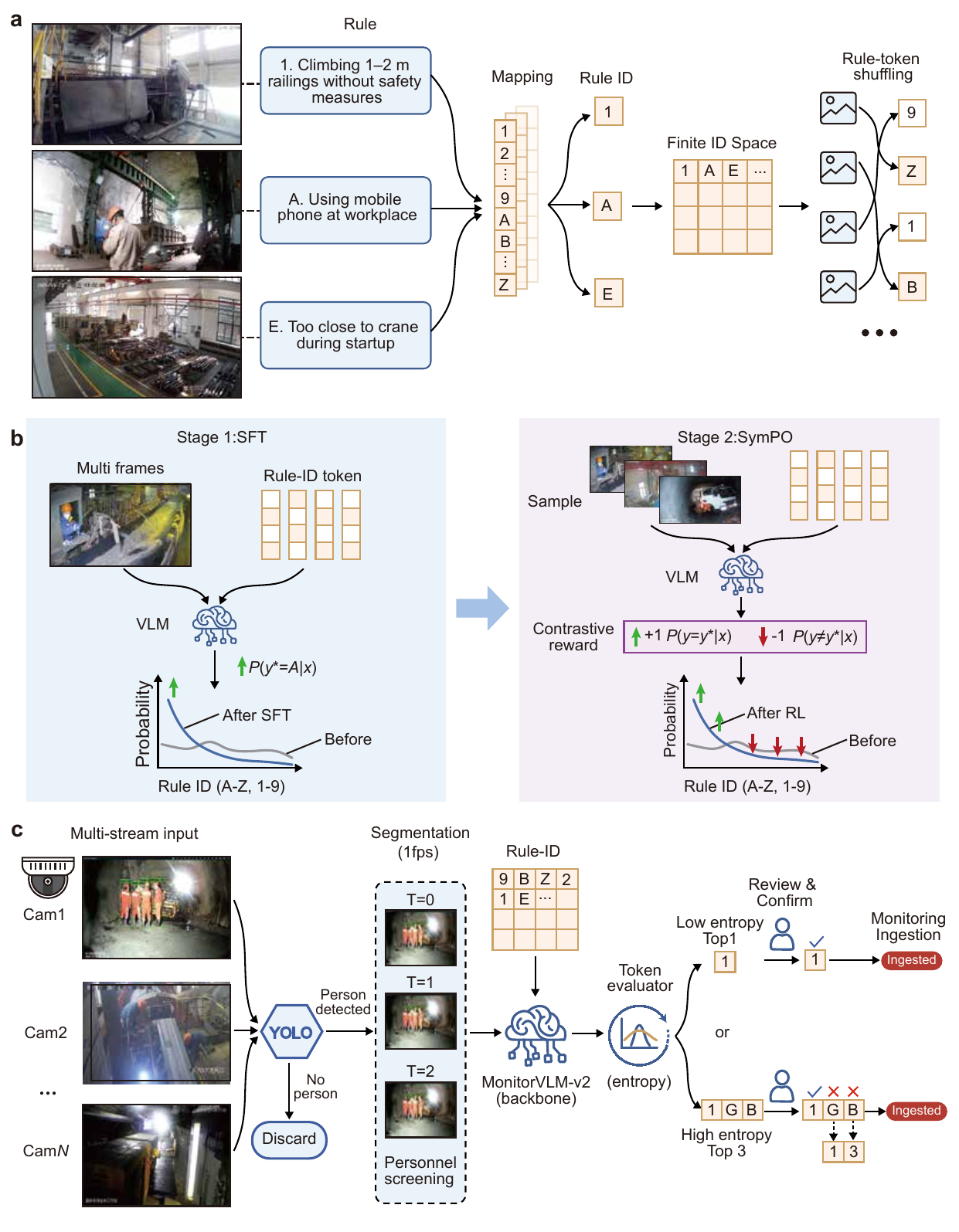}
\end{figure}

\begin{center}
    \captionsetup{type=figure, font=footnotesize}   % ← 加 font=small
  \caption{\bodyfigurelabel{fig.ACD1}
    \textbf{Probabilistic decision architecture of MonitorVLM-v2.}
a, Construction of a bounded symbolic decision space. Safety observations are 
mapped to a finite set of rule-ID tokens corresponding to predefined safety regulations. 
Rule-token shuffling (RTS) periodically permutes the rule-to-token mapping during 
training, reducing reliance on fixed token identities and encouraging predictions to be 
grounded in visual-regulatory correspondence rather than token-index shortcuts.
b, Two-stage contrastive optimization in the symbolic decision space. In Stage 1, 
supervised fine-tuning (SFT) shifts the probability mass toward the correct rule-ID 
(shown as the rising curve after SFT). In Stage 2, SymPO further concentrates 
probability on the correct hypothesis while actively suppressing competing rule-IDs 
(downward arrows on incorrect candidates after RL), improving rule-level discrimination 
over the SFT baseline.
c, \textbf{Uncertainty-guided deployment under multistream surveillance.}
Concurrent surveillance streams are first screened by a lightweight YOLO-based personnel sentinel 
at 1 fps; human-containing segments are forwarded to MonitorVLM-v2 for single-step 
rule-ID decoding. Prediction entropy routes outputs along two paths: low-entropy 
predictions are submitted for Top-1 human confirmation, whereas high-entropy predictions 
indicating perceptual ambiguity are escalated to compact Top-3 candidate sets for expert 
inspection.}
\end{center}

\section*{\LARGE Results}
We validated MonitorVLM-v2 through a staged evaluation progressing from controlled model analysis to prospective field deployment. Experiments assessed the extent to which CoT-style supervision could be compressed into single-step symbolic decisions without substantially compromising regulatory discrimination; the contribution of contrastive policy optimization to decision-boundary separation; the diagnostic utility of prediction entropy; and finally the complete pipeline across 10 concurrent streams in an operational underground mining facility over four months.

The initial training set contained 8,002 violation events and 5,077 nonviolation events, which covered 35 mining safety regulations. The test set comprised 862 events (732 violations and 130 nonviolations). All events were extracted from surveillance footage recorded by the 100 fixed high-definition cameras installed across the operational areas of the target underground mining facility. Violation clips were annotated by the site's trained safety inspectors, who verified each event against the corresponding regulatory requirement and assigned an exact rule-ID label from the 35-class regulatory vocabulary.
We benchmarked MonitorVLM-v2 against proprietary multimodal models\cite{yao2025efficient,comanici2025gemini,choi2025comparison}, each evaluated zero-shot, and four open-source VLM backbones\cite{bai2025qwen3,zhu2025internvl3,gemmateam2025gemma3technicalreport,liu2023visual,li2024llava} evaluated both before and after domain adaptation.
 We adopted exact rule-ID accuracy as the primary endpoint: an identification was scored as correct only if the predicted regulation matched the ground truth exactly, preventing misclassifications between rules from inflating reported performance. We additionally report recall, precision, and F1 score, as missed violations carry greater operational risk than false alarms.

\noindent\textbf{Reasoning compression preserves regulatory discrimination.}
A central question for MonitorVLM-v2 is whether explicit online CoT generation is necessary for bounded regulatory tasks. We addressed this by fine-tuning the same base model backbones under three supervision regimes (Fig.~\ref{fig.ACD2}a): CoT supervision, producing full natural-language reasoning traces of up to 8,192 tokens; rule-ID supervision without RTS, in which each regulation is mapped to a fixed token throughout training; and rule-ID supervision with RTS, 
in which the rule-to-token correspondence is permuted at the start of each epoch. Rule-ID with RTS retained most of the discriminative capacity of CoT supervision, incurring an average accuracy reduction of only 4.66\% across all four backbones (Fig.~\ref{fig.ACD2}b), while achieving a 19.45-fold reduction in end-to-end inference latency and satisfying the sub-second throughput requirement of real-time multistream surveillance (Fig.~\ref{fig.ACD2}c).

\clearpage
\begin{figure}[!h] 
    \centering
    \includegraphics[width=0.8\textwidth]{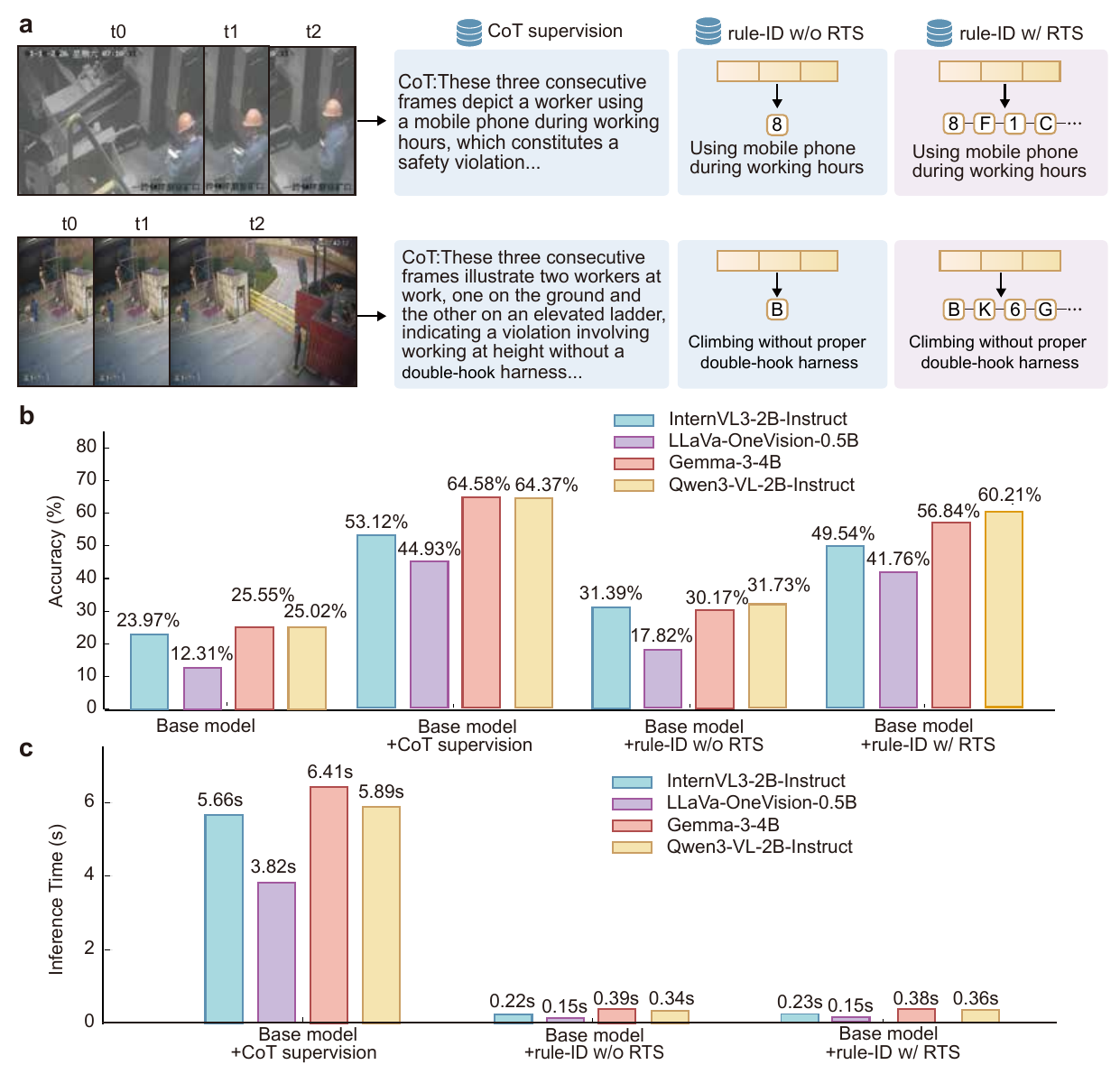}
\end{figure}
\begin{center}
    \captionsetup{type=figure, font=footnotesize}   % ← 加 font=small
    \caption{\bodyfigurelabel{fig.ACD2}
    \textbf{Reasoning compression preserves regulatory discrimination while collapsing inference latency.}
a, Three supervision regimes for evaluating CoT compression, illustrated with representative examples. CoT supervision produces full natural-language reasoning traces of up to 8,192 tokens. Rule-ID supervision without RTS maps each regulation to a fixed token throughout training. Rule-ID supervision with RTS periodically reshuffles the rule-to-token correspondence to prevent token-identity shortcuts.
b, Symbolic supervision retains regulatory discrimination across four VLM backbones. Rule-ID with RTS incurs an average accuracy reduction of only 4.66 percentage 
points relative to full CoT supervision, while RTS consistently outperforms fixed 
rule-ID mapping by forcing visual-semantic grounding. Qwen3-VL-2B-Instruct achieves 
the highest accuracy of 60.21\% under rule-ID with RTS.
c, Single-token prediction collapses inference latency. Replacing autoregressive 
CoT generation with bounded rule-ID prediction reduces online decoding time by 19.45-fold (from 3.82-6.41 s to 0.15-0.39 s per three-frame segment), enabling subsecond throughput and satisfying the real-time requirement of multistream deployment.
    }
\end{center}

A direct comparison of rule-ID supervision with and without RTS reveals that conditional inference over the rule-semantic--symbol mapping is essential for robust symbolic compression. Despite both settings eliminating CoT generation, rule-ID without RTS consistently underperformed its with-RTS counterpart, indicating that a static mapping causes the model to exploit fixed token-index associations rather than reasoning from visual and regulatory evidence. By periodically reshuffling the mapping, RTS compels the model to condition its predictions on the active rule-semantic assignment, thereby acquiring dynamic rule-table adaptation capacity and enhancing robustness across variable regulatory configurations and visual conditions. Among the evaluated backbones, Qwen3-2B-basic achieved the highest accuracy of 60.21\% under rule-ID with RTS, exceeding LLaVA-0.5B-basic by 18.45 percentage points, and was selected as the backbone for subsequent policy optimization (Table~\ref{tab:benchmark}). 

\noindent\textbf{Two-stage optimization sharpens decision boundaries.}
\label{Two-stage fine-tuning results}
Building on the Qwen3-2B-basic SFT checkpoint,, we further optimized the model using reinforcement learning to sharpen decision boundaries in the finite rule-ID space. Supervised likelihood maximization reinforces the ground-truth rule-ID but does not explicitly suppress competing alternatives, leaving the model vulnerable to visually ambiguous regulatory hypotheses. We therefore compared SymPO against two 
group-based reinforcement learning objectives, GRPO and DAPO\cite{yu2025dapo,liu2026stapo,guo2025deepseek,jacobs2023deepspeed,schulman2017proximal}. As shown in Fig.~\ref{fig.curse}a, SymPO converged rapidly to an 
accuracy plateau of 69.84\%, whereas the accuracies of GRPO and DAPO 
saturated near 65.2\%. This improvement was driven primarily by a substantial increase in recall, from 58.14\% for the SFT backbone to 74.73\% for MonitorVLM-v2, indicating that more true violations were successfully recovered under the strict rule-ID protocol. Its precision was slightly lower than that of the most conservative SFT backbone, but remained high at 87.94\%. This trade-off reflects the safety-oriented objective of the system: recovering more true violations and reducing missed detections, rather than maximizing alarm precision alone (Table~\ref{tab:benchmark}).

Examination of token-level probability transitions reveals the mechanism underlying this improvement (Fig.~\ref{fig.curse}b). SymPO produced a markedly asymmetric probability shift: it concentrated probability mass near 1.0 for correctly classified samples while, critically, suppressing high-confidence distractor rule-IDs in misclassified cases. This asymmetry indicates that SymPO reshapes the finite symbolic decision space by penalizing competing hypotheses, sharpening boundaries between visually adjacent regulatory violations. The resulting MonitorVLM-v2 outperformed all fine-tuned open-source baselines and surpassed zero-shot proprietary models including Claude Sonnet 4.6 and GPT-5.2 under the reported evaluation protocol (Table~\ref{tab:benchmark}).

\begin{figure}[!h] 
    \centering
    \includegraphics[width=1.0\textwidth]{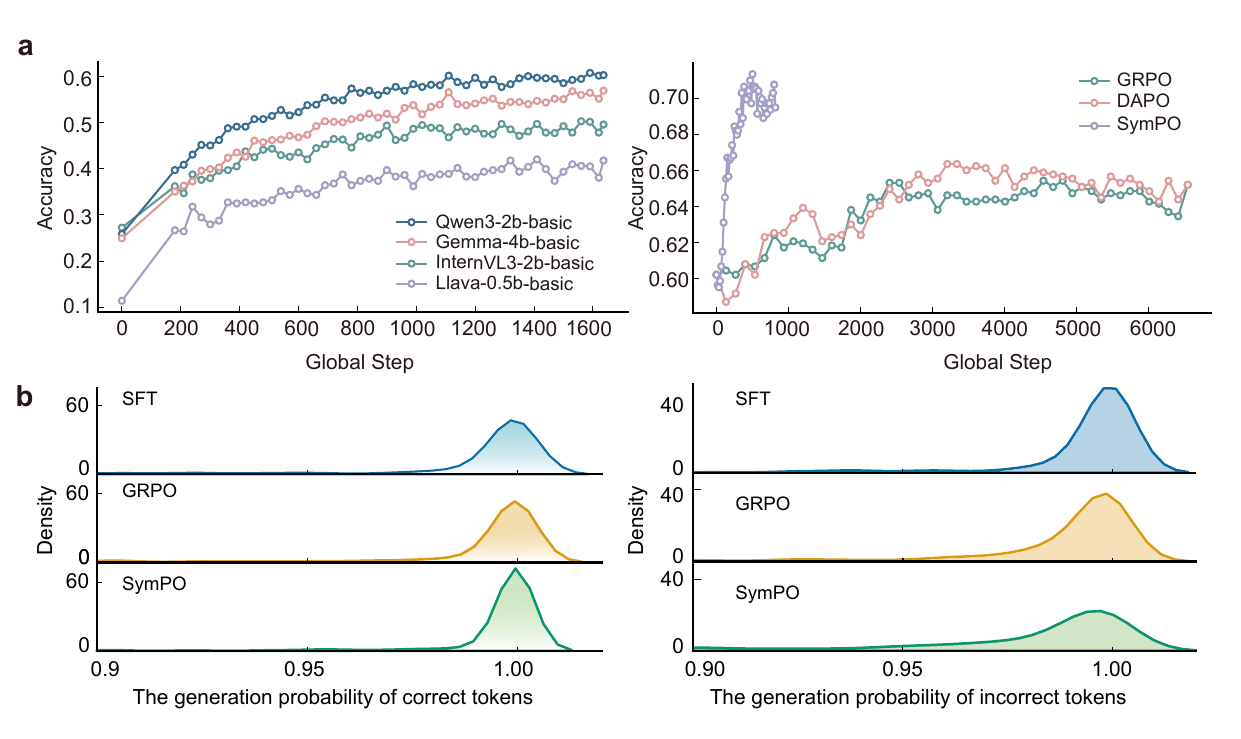}
\end{figure}
\begin{center}
    \captionsetup{type=figure, font=footnotesize}   % ← 加 font=small
    \caption{\bodyfigurelabel{fig.curse}
    \textbf{Two-stage optimization dynamics and token-level probability analysis.}
a, Training convergence curves. Left: SFT accuracy curves for four backbone 
architectures over 1,600 training steps, showing progressive domain alignment on the 
rule-ID task. Right: RL-stage accuracy curves comparing GRPO, DAPO, and SymPO 
initialized from the Qwen3-2B-basic SFT checkpoint over 6,000 steps. SymPO converges 
to 69.84\%, surpassing GRPO (65.20\%) and DAPO (65.22\%).
b, Token-level generation probability distributions for correctly classified 
(left) and incorrectly classified (right) samples across SFT, GRPO, and SymPO. For 
correct predictions, all three methods concentrate probability near 1.0. For incorrect predictions, SymPO uniquely suppresses high-confidence distractor rule-IDs, producing a broader, lower-probability distribution compared with SFT and GRPO, which indicates that SymPO reshapes the decision space by penalizing competing hypotheses rather than only reinforcing the correct one.
    }
\end{center}

\clearpage
\begin{table}[!htbp]
\centering
\caption{Performance comparison across multimodal models. Metrics were computed under a strict rule-ID protocol: an accuracy was counted as correct only if the predicted regulation exactly matched the ground truth (nonviolation was treated as a class). Furthermore, 95\% confidence intervals ($\pm$) were estimated via bootstrap resampling (1,000 iterations) on the held-out test set ($n=862$).}
\label{tab:benchmark}
\resizebox{\textwidth}{!}{%
\begin{tabular}{lcccc}
\toprule
\textbf{Algorithm} & \textbf{Accuracy} & \textbf{Precision} & \textbf{Recall} & \textbf{F1 score} \\
\midrule
\multicolumn{5}{l}{\textit{Proprietary models (no fine-tuning)$^\dagger$}} \\
\midrule
Claude Sonnet 4.6
  & $56.73\%_{\pm3.31}$
  & $87.47\%_{\pm2.87}$
  & $57.24\%_{\pm3.65}$
  & $69.20\%_{\pm2.47}$ \\
GPT-5.2
  & $32.13\%_{\pm2.19}$
  & $84.51\%_{\pm1.80}$
  & $24.59\%_{\pm1.22}$
  & $38.10\%_{\pm2.03}$ \\
Gemini-3-Flash
  & $40.38\%_{\pm3.26}$
  & $84.42\%_{\pm3.07}$
  & $31.86\%_{\pm3.23}$
  & $46.26\%_{\pm3.41}$ \\
\midrule
\multicolumn{5}{l}{\textit{Open-source models (zero-shot, original checkpoints)$^\ddagger$}} \\
\midrule
InternVL3-2B-Instruct
  & $27.26\%_{\pm2.98}$
  & $74.42\%_{\pm3.49}$
  & $21.86\%_{\pm2.93}$
  & $33.79\%_{\pm3.58}$ \\
LLaVA-OneVision-0.5B
  & $11.46\%_{\pm2.14}$
  & $64.86\%_{\pm4.17}$
  & $16.39\%_{\pm2.61}$
  & $26.17\%_{\pm3.82}$ \\
Gemma-3-4B
  & $24.94\%_{\pm2.91}$
  & $70.73\%_{\pm3.68}$
  & $19.81\%_{\pm2.80}$
  & $30.95\%_{\pm3.71}$ \\
Qwen3-VL-2B-Instruct
  & $25.87\%_{\pm2.94}$
  & $72.04\%_{\pm3.59}$
  & $20.77\%_{\pm2.85}$
  & $32.24\%_{\pm3.66}$ \\
\midrule
\multicolumn{5}{l}{\textit{Open-source models (fine-tuned on the domain dataset)$^\S$}} \\
\midrule
InternVL3-2B-basic
  & $49.54\%_{\pm3.37}$
  & $89.18\%_{\pm2.73}$
  & $46.17\%_{\pm3.68}$
  & $60.84\%_{\pm3.43}$ \\
LLaVA-0.5B-basic
  & $41.76\%_{\pm3.30}$
  & $83.82\%_{\pm3.17}$
  & $38.93\%_{\pm3.55}$
  & $53.17\%_{\pm3.62}$ \\
Gemma-4B-basic
  & $56.84\%_{\pm1.87}$
  & $91.86\%_{\pm2.38}$
  & $53.96\%_{\pm1.52}$
  & $67.98\%_{\pm1.64}$ \\
Qwen3-2B-basic
  & $60.21\%_{\pm2.13}$
  & $93.32\%_{\pm2.20}$
  & $58.14\%_{\pm3.12}$
  & $71.64\%_{\pm1.86}$ \\
Qwen3-2B-basic + GRPO
  & $65.20\%_{\pm1.97}$
  & $91.70\%_{\pm2.29}$
  & $64.89\%_{\pm3.40}$
  & $76.00\%_{\pm2.64}$ \\
Qwen3-2B-basic + DAPO
  & $65.22\%_{\pm3.13}$
  & $91.73\%_{\pm2.34}$
  & $64.85\%_{\pm3.28}$
  & $75.98\%_{\pm2.57}$ \\
\midrule
\textbf{MonitorVLM-v2 (ours)}
  & $\mathbf{69.84\%_{\pm2.07}}$
  & ${87.94\%_{\pm2.31}}$
  & $\mathbf{74.73\%_{\pm3.22}}$
  & $\mathbf{80.80\%_{\pm2.28}}$ \\
\bottomrule
\end{tabular}}
\smallskip
\raggedright\footnotesize
$^\dagger$ Evaluated through official APIs without domain-specific fine-tuning. These models are included as zero-shot diagnostic baselines and are not used to establish direct superiority over domain-adapted models.
$^\ddagger$~Original open-source checkpoints evaluated with the same rule-ID
prompt but without any domain fine-tuning.\\
$^\S$~``-basic'' denotes the corresponding backbone after Stage-1 SFT on the RTS domain dataset (Methods).
MonitorVLM-v2 corresponds to Qwen3-2B-basic further optimized with SymPO.
\end{table}

\noindent\textbf{Entropy-guided triage identifies ambiguous failure modes.}
The two-stage optimization substantially improved regulatory discrimination, yet the remaining errors were not random noise but formed a structured pattern concentrated in visually and contextually challenging scenarios. Correct and incorrect identifications were unevenly distributed across the 35 safety regulations (TP 547, FN 185, FP 75, and TN 55), and false negatives disproportionately occurred in rules that rely on weak visual cues, fine-grained posture recognition, or complex contextual reasoning (Fig.~\ref{fig.failure}a). To investigate the origin of these errors, we examined representative failure cases and their attention patterns (Fig.~\ref{fig.failure}b). Misclassifications consistently arose from perceptual ambiguity, including low illumination, strong glare, severe occlusion, and large camera-to-subject distances. Grad-CAM\cite{selvaraju2017grad} visualizations reveal that in these scenarios, model attention was often diffuse or misaligned with the regions critical for correct regulatory judgement. Notably, even with explicit textual CoT generation removed to reduce latency, the system preserved spatial explainability through the final cross-attention maps extracted from the visual encoder.

\clearpage
\begin{figure}[!h] 
    \centering
    \includegraphics[width=0.75\textwidth]{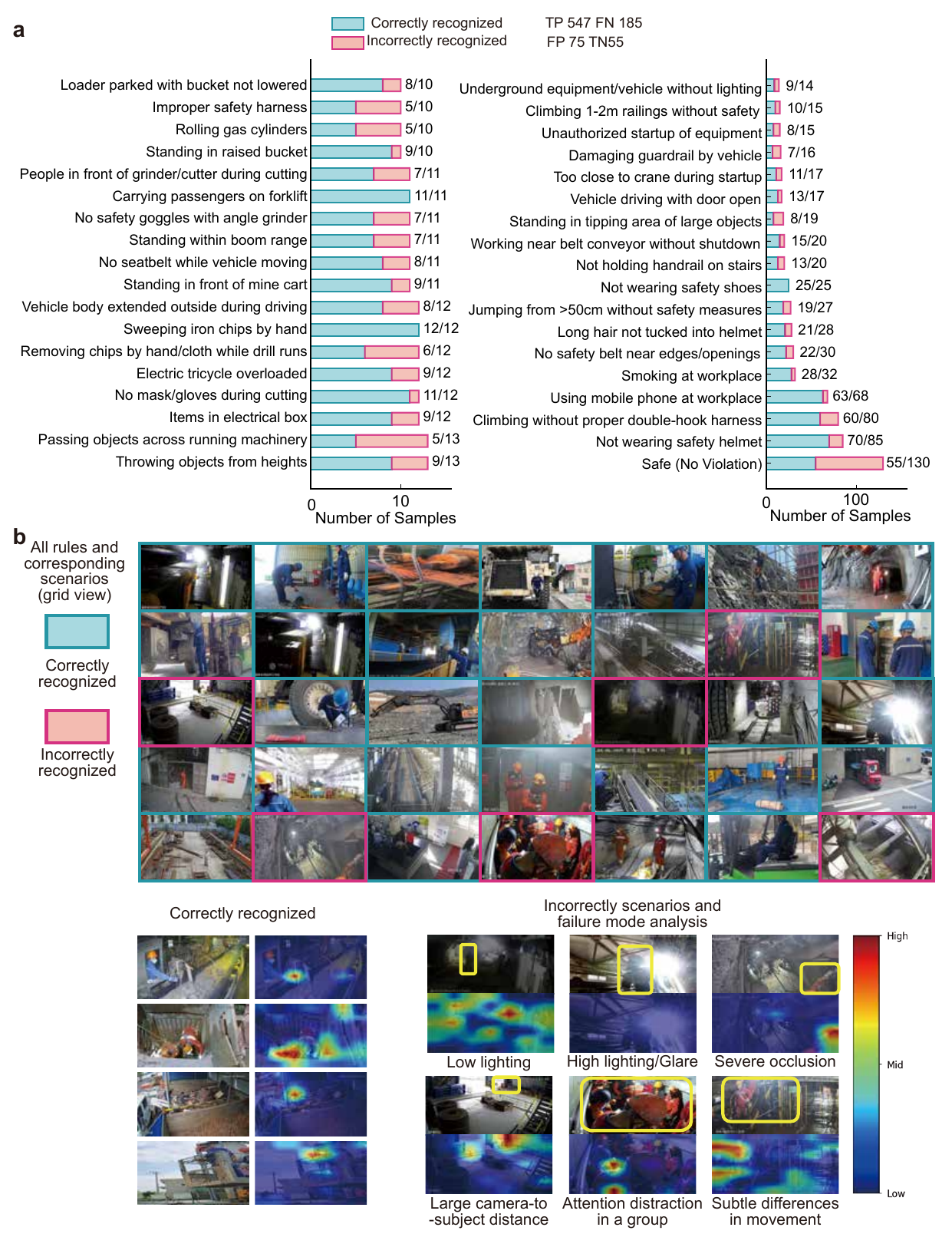}
\end{figure}
\begin{center}
    \captionsetup{type=figure, font=footnotesize}   % ← 加 font=small
    \caption{\bodyfigurelabel{fig.failure}
\textbf{Structured error distribution and perceptual failure analysis.}
\textbf{a}, \textbf{Regulatory discrimination profile.} Distribution of correct (TP, TN) and incorrect (FP, FN) predictions across 35 safety regulations (TP 547, FN 185, FP 75, 
TN 55). False negatives are disproportionately concentrated in regulations requiring weak visual cues, fine-grained posture recognition, or complex contextual reasoning, 
indicating that residual errors are structured rather than random.
\textbf{b}, \textbf{Visual failure modes and attention alignment.} Left: representative 
correctly recognized cases with their Grad-CAM attention maps, where high-activation 
regions (red) align with task-relevant areas. Right: six perceptual failure categories 
with corresponding Grad-CAM visualizations, including low lighting, high lighting and 
glare, severe occlusion, large camera-to-subject distance, attention distraction within 
a group, and subtle differences in movement. In each failure case, model attention is 
diffuse or misaligned with the regions critical for correct regulatory judgement, 
explaining the systematic concentration of errors in these visual conditions.
}
\end{center}

The structured nature of these failures suggests that situational ambiguity can be explicitly measured and operationalized as a routing signal. Misclassified samples consistently exhibited higher Shannon entropy $H(I,\mathcal{R})$ than correctly classified ones across all optimization regimes, and SymPO produced the clearest entropy separation between correct and incorrect predictions (Fig.~\ref{fig.entropy}a), indicating that contrastive optimization strengthens the diagnostic utility of entropy as an uncertainty measure.

We therefore implemented an entropy-governed triage mechanism routing low-entropy segments to a Top-1 confirmation path and elevating high-entropy cases to a compact Top-3 candidate set for expert review. As shown in Fig.~\ref{fig.entropy}b, at an operational threshold of $\tau=10^{-3}$, the Top-1 path correctly resolved 602 cases, and the Top-3 triage branch recovered an additional 247 difficult cases, which corresponded to 28.65\% of the test set. By operationalizing uncertainty as a measurable routing signal, the framework allocates human cognitive effort to the most ambiguous scenarios and preserves the high-throughput advantages of autonomous symbolic inference.
\clearpage
\begin{figure}[!h] 
    \centering
    \includegraphics[width=0.85\textwidth]{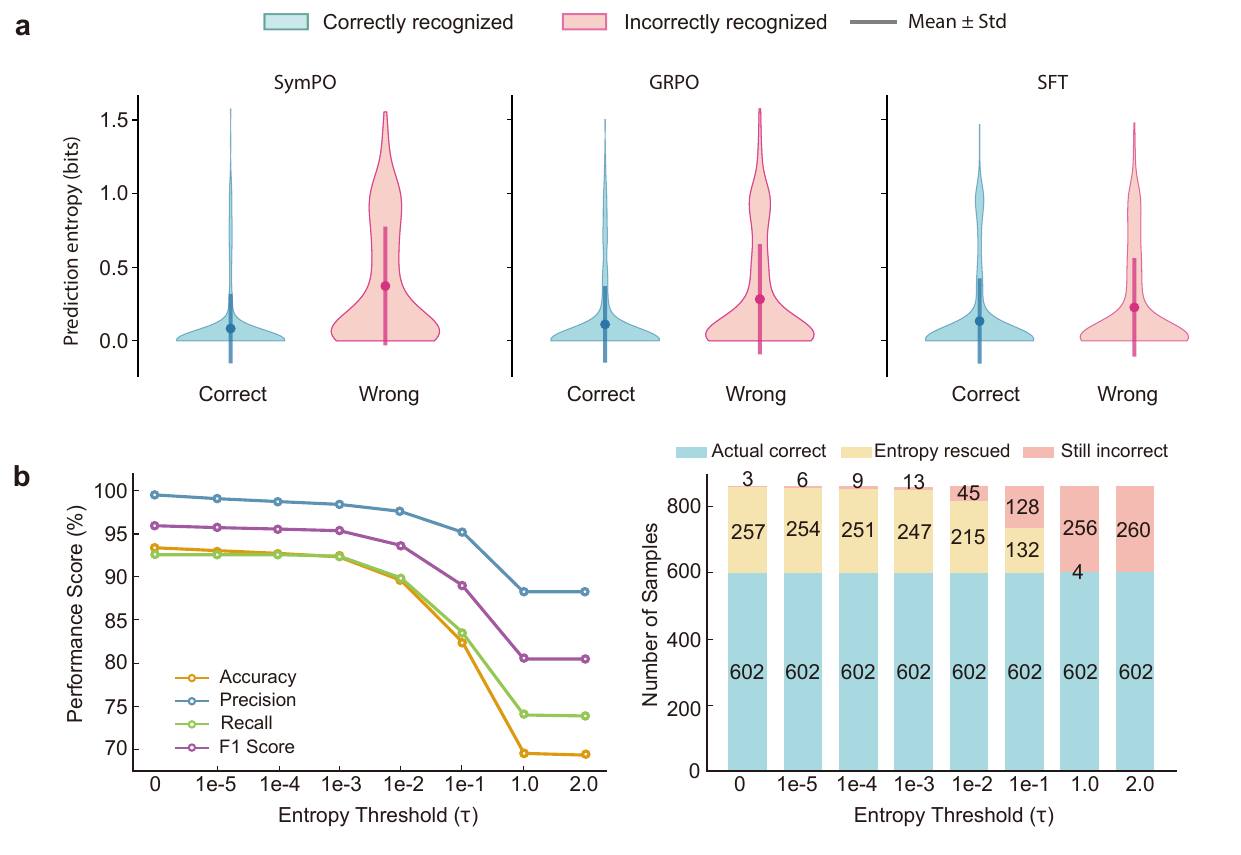}
\end{figure}
\begin{center}
    \captionsetup{type=figure, font=footnotesize}   % ← 加 font=small
    \caption{\bodyfigurelabel{fig.entropy}
\textbf{Uncertainty-guided routing and human--AI triage.}
\textbf{a}, Entropy distributions across optimization regimes. Violin plots show the 
distribution of prediction entropy (width proportional to density; centre mark 
denotes mean) for correctly and incorrectly classified samples under SFT, GRPO, and 
SymPO. SymPO produces the clearest separation between low-entropy correct predictions 
and high-entropy errors, indicating that contrastive optimization strengthens the 
diagnostic utility of entropy as an uncertainty signal.
\textbf{b}, Triage performance across entropy thresholds. Left: accuracy, precision, 
recall, and F1 score as a function of entropy threshold $\tau$. Right: stacked bars 
showing, at each threshold, the number of cases correctly resolved through the 
low-entropy Top-1 confirmation path (blue), additional cases recovered through the 
high-entropy Top-3 expert review path (yellow), and cases remaining incorrect (pink). 
At the operational deployment threshold $\tau = 10^{-3}$ (vertical dashed line), the 
Top-1 path correctly resolves 602 cases and the Top-3 branch recovers an additional 
247 cases (28.65\% of the test set), demonstrating a controllable trade-off between 
autonomous throughput and human review coverage.
}
\end{center}

\noindent\textbf{Real-world deployment and operational efficiency.}
In practice, the rule-ID formulation also simplifies dataset maintenance relative to CoT-style annotation, because safety inspectors need only verify and correct bounded regulatory labels rather than revise free-form reasoning traces. This property enabled an efficient closed-loop adaptation phase before full deployment, using historical and pilot surveillance clips collected from the target mining facility. In each iteration, inspectors reviewed high-entropy predictions and representative failure cases, corrected labels, and returned 500 newly verified samples to the training set per round. After four iterations, the detection accuracy increased from 69.84\% to 77.21\%, whereas that of the static model remained near 69.90\% (Fig.~\ref{fig.online}a). These results indicate that the feedback loop helped adapt MonitorVLM-v2 to the target operational distribution.
After this adaptation phase, the finalized model was evaluated prospectively for four months in an operational mining facility. The system monitored 10 high-definition video streams from 10 fixed cameras concurrently using a two-stage edge pipeline, providing uninterrupted 24-hour surveillance coverage. Under the site's existing inspection workflow, two trained safety inspectors monitored the same 10 camera feeds in rotating 12-hour shifts, yielding 18 hours of daily manual coverage; the remaining 6 hours overnight fell outside the routine inspection window. A lightweight YOLOv11\cite{khanam2024yolov11} sentinel first screened frames for human presence at 1~fps, and only human-containing segments were forwarded to MonitorVLM-v2 for rule-level reasoning.

During the four-month deployment, MonitorVLM-v2 identified 89 confirmed violations under continuous 24-hour monitoring, whereas 32 violations were identified via routine manual inspection (18 hours of daily coverage by two inspectors in 12-hour shifts) under the existing workflow of the site (Fig.~\ref{fig.online}b), which corresponded to 2.78 times as many confirmed detections as the manual inspection workflow recorded over the same period. Among the 96 confirmed violations reviewed during the trial, 64 were identified only by MonitorVLM-v2, 25 were detected by both the system and human operators, and 7 were captured only through manual inspection. These results suggest that the system can complement human inspectors by identifying violations that may be missed under routine monitoring. Human operators remain essential for verification and final accountability, whereas MonitorVLM-v2 functions as a persistent monitoring engine that reduces missed events caused by limited inspection frequency, attention fatigue, and discontinuous coverage.

A detailed audit of the deployment results (Fig.~\ref{fig.online}c) reveals the current boundaries of the system. During the trial, seven incorrect identifications occurred across four categories: Driving with door open was misclassified because of the model's difficulty in distinguishing subtle vehicle motion from static idling; not holding handrail on stairs and climbing 1--2 m railings were affected by large camera-to-subject distances and perspective distortion; and working near the belt conveyor was occasionally misidentified as a fall-protection violation because of height-related perceptual ambiguity. These findings highlight the perceptually ambiguous scenarios that remain challenging and indicate potential directions for improving temporal modelling and viewpoint adaptation.

\clearpage
\begin{figure}[!h] 
    \centering
    \includegraphics[width=0.85\textwidth]{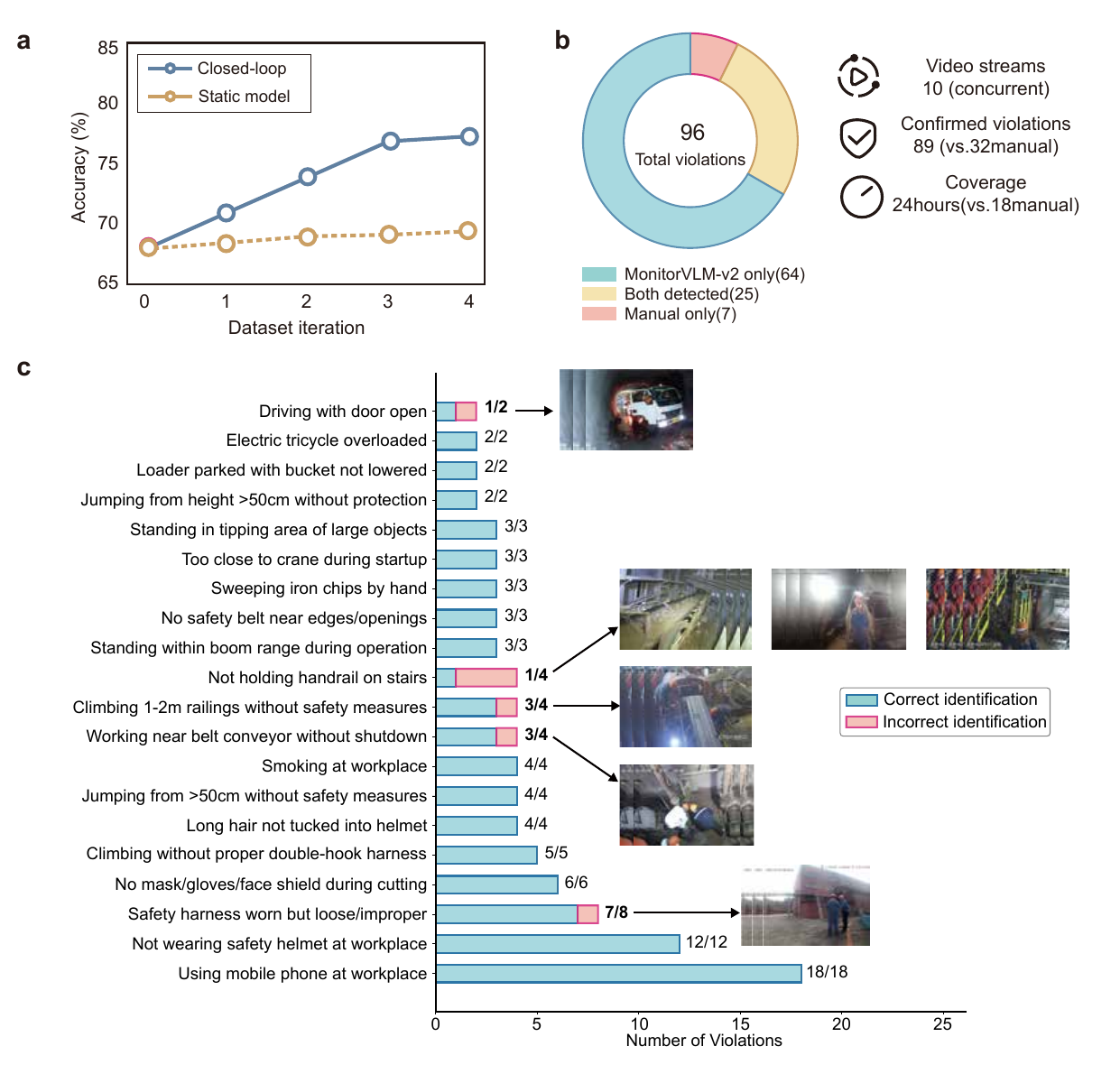}
\end{figure}
\begin{center}
    \captionsetup{type=figure, font=footnotesize}   % ← 加 font=small
    \caption{\bodyfigurelabel{fig.online}
  \textbf{Real-world deployment and operational analysis.}
   \textbf{a}, Closed-loop adaptation. Detection accuracy over four dataset iterations for 
the feedback-updated model (closed-loop) and a static baseline. Each iteration 
incorporates 500 newly labelled error cases identified from high-entropy predictions at 
the target site. The closed-loop model improves progressively from 69.84\% to 77.21\% 
over four iterations, whereas the static model remains near 69.90\%, confirming that 
bounded rule-ID annotation enables tractable iterative adaptation.
    \textbf{b}, Operational detection summary. Donut chart showing the overlap between 
MonitorVLM-v2 and routine manual inspection across 96 confirmed violations reviewed 
during the four-month trial: 64 identified exclusively by MonitorVLM-v2, 25 identified 
by both, and 7 identified exclusively by manual inspection. Right panel summarises 
deployment statistics: 10 concurrent video streams, 89 confirmed violations detected 
under continuous 24-hour monitoring versus 32 under the existing manual workflow.
   \textbf{c}, Regulatory audit and failure cases. Horizontal bars show the number of 
correct (teal) and incorrect (pink) identifications for each safety regulation observed 
during deployment. High-salience violations such as mobile phone use (18/18) and missing 
safety helmet (12/12) were detected with perfect accuracy, whereas errors were 
concentrated in visually ambiguous categories including handrail use, railing climbing, 
and distance-distorted actions. Representative images of both correctly identified 
violations and misclassified cases are shown alongside the corresponding regulation bars.}
\end{center}

\section*{\LARGE Discussion}
We presented MonitorVLM-v2, a deployment-oriented framework for real-time safety violation detection in safety-critical industrial environments. By reformulating multimodal reasoning as bounded symbolic decision-making, MonitorVLM-v2 integrates rule-ID prediction, contrastive symbolic policy optimization and entropy-guided human--AI triage into a scalable monitoring architecture. The results suggest that, for rule-governed industrial surveillance, the key deployment challenge is not the production of longer natural-language rationales but the conversion of visual evidence and regulatory knowledge into fast, auditable and uncertainty-aware decisions.

A principal finding of this work is that open-ended autoregressive CoT reasoning is not a functional necessity for bounded regulatory tasks. Replacing free-form generation with symbolic prediction over a finite rule space preserved rule-level discrimination while reducing online decoding latency by 19.45-fold. These results indicate that, in safety-critical domains, the functional role of reasoning can be captured by calibrated decision distributions rather than explicit natural-language traces. Such compression reduces the decoding complexity from $O(T)$ to $O(1)$ with respect to the output length and provides a more predictable interface for auditable industrial deployment.

SymPO further addresses the optimization problem introduced by this finite symbolic formulation. Standard supervised fine-tuning increases the likelihood of the ground-truth rule-ID but does not explicitly suppress visually plausible distractors\cite{ouyang2022training,rafailov2023direct}. SymPO instead treats symbolic prediction as a contrastive policy-learning problem and assigns positive rewards to the correct regulatory hypothesis and negative rewards to competing alternatives. This objective improves separation within the finite decision space and sharpens the boundaries between visually adjacent safety violations, thus making the symbolic output distribution more suitable for deployment.

A second key insight is that residual errors should be treated as structured signals rather than random noise. The remaining failures are concentrated in visually ambiguous conditions, including occlusion, degraded illumination, long camera-to-subject distance and subtle action differences. By quantifying predictive uncertainty through Shannon entropy, MonitorVLM-v2 converts such ambiguity into a measurable routing signal. High-entropy cases can therefore be directed to expert review, and low-entropy cases can be handled through rapid Top-1 confirmation. This design establishes a practical interface between autonomous symbolic inference and human supervisory judgement, thus preserving auditability and reducing unnecessary expert workload.

The four-month prospective deployment provides real-world evidence for the practical value of this architecture. To our knowledge, this is the first VLM-based safety monitoring system evaluated under continuous 24-hour operation over multiple months, establishing a longitudinal benchmark for sustained industrial reliability. Continuous AI-assisted monitoring identified 2.78 times as many confirmed violations as the parallel routine manual inspection workflow, addressing the fundamental limitations of discontinuous coverage and attention fatigue inherent in human-only surveillance. Importantly, the system is designed as a complement to, not a replacement for, human oversight: human operators remain essential for final verification, contextual judgement, and legal accountability. The closed-loop adaptation mechanism further demonstrates that the bounded annotation space of rule-ID labels enables tractable iterative improvement on site-specific failure cases, a property that would be substantially more difficult to realize with free-form CoT supervision.

Several limitations define directions for future development. The most persistent failure modes, including perspective distortion, extreme occlusion, and temporally extended behaviours, reflect the constraints of static spatial reasoning applied to dynamic three-dimensional environments. Future work should investigate stronger temporal modelling across extended observation windows, multi-view fusion to resolve occlusions, and viewpoint-invariant representations. Uncertainty calibration beyond threshold-based triage, for instance via conformal prediction or Bayesian approaches, could provide tighter statistical guarantees on missed-violation rates at specified confidence levels. More broadly, the reasoning-to-decision compression paradigm proposed here offers a principled pathway for adapting VLMs from open-ended multimodal assistants into reliable, accountable decision modules for diverse rule-governed industrial domains.

\section*{\LARGE Methods}
\noindent\textbf{Discrete symbolic mapping and vocabulary construction.}
MonitorVLM-v2 formulates safety monitoring as prediction over a bounded regulatory vocabulary. For each input segment \(I=\{I_1,I_2,I_3\}\), we define a symbolic decision space \(\mathcal{Y}=\{y_0,y_1,\ldots,y_K\}\), where \(K=35\) corresponds to the number of safety regulations and \(y_0\) denotes the nonviolation class. The model output is constrained to this vocabulary using a logit-masking layer to ensure that the distribution \(\pi_\theta(y\mid I,\mathcal{R})\) is defined only over valid regulatory hypotheses.

We let \(\phi:\mathcal{R}\rightarrow\mathcal{Y}\) denote the mapping from safety regulations to rule-ID tokens. During training, \(\phi\) is reshuffled at the start of each epoch, and the semantic association between each visual segment and its corresponding regulation is preserved. This reshuffling compels the model to perform conditional inference over the current rule-semantic--symbol assignment, endowing it with dynamic rule-table adaptation capacity and ensuring that predictions depend on the regulatory semantics of the active mapping rather than on static token-index associations. During deployment, \(\phi\) is fixed to a deterministic dictionary to ensure that the rule-ID outputs can be interpreted unambiguously by downstream monitoring systems.

\noindent\textbf{Two-stage optimization.}
MonitorVLM-v2 is trained in two stages. In Stage~1, SFT is performed using cross-entropy loss on the RTS training set. In Stage~2, SymPO is initialized from the SFT checkpoint. For each prompt, the model produces a single symbolic sample from the finite rule-ID space. The sampled token receives an absolute contrastive reward, with +1 assigned to the correct rule-ID and -1 assigned to all incorrect rule-IDs. Compared with the group-relative baselines GRPO and DAPO, which use \(G=8\) samples per prompt, SymPO uses one sample per prompt and reduces the per-step rollout cost by 87.5\% in our implementation.

\noindent\textbf{Entropy-based triage calibration.}
For deployment, the Shannon entropy \(H(I,\mathcal{R})\) of the symbolic output distribution is used as an uncertainty score. The threshold \(\tau\) is selected by grid search on the validation set to balance expert review and expected recall. Segments with ($H < \tau$) enter the Top-1 confirmation path, and segments with ($H \ge \tau$) are presented as compact Top-3 candidates for expert inspection. In deployment, we set $\tau=10^{-3}$ and use Top-3 review for high-entropy cases. Under this dual-pathway architecture, every prediction requires human confirmation before being recorded, ensuring a complete and auditable human review trail with no autonomous final judgements.

\noindent\textbf{Deployment pipeline and feedback buffer.}
The deployed system uses a sentinel--reasoner pipeline. A YOLO--v11 personnel sentinel screens video streams at 1~fps. Frames without detected personnel are discarded, and human-containing segments are forwarded to MonitorVLM-v2. For each detected segment, three frames are sampled from a 3-second observation window and used for rule-ID prediction. Verified violations and corrected false positives from the human review interface are stored in a hard-example buffer. These samples are incorporated into subsequent adaptation rounds to improve performance on site-specific failure cases.

\noindent\textbf{Experimental setup and hyperparameters.}
All experiments were conducted on 8 NVIDIA A40 GPUs using DeepSpeed ZeRO-3\cite{jacobs2023deepspeed} optimization and BF16 precision. For SFT, VLM backbones were trained for 2 epochs with a per-device batch size of 1 and gradient accumulation to obtain an effective total batch size of 16. The learning rate was \(1\times10^{-4}\) with cosine decay, the maximum sequence length was 8,192 tokens, and Low-Rank Adaptation (LoRA)\cite{hu2022lora} was applied with rank $r=16$ and $\alpha=32$.
For SymPO, training was conducted for 1 epoch from the SFT checkpoint, with a per-device batch size of 4, gradient accumulation of 1 and a learning rate of \(1\times10^{-6}\). Symbolic sampling used temperature 0.7 and top-$p$ 0.85. GRPO and DAPO baselines used ($G=8$) samples per prompt under the same symbolic rule-ID evaluation protocol. Data loading used 4 workers per GPU.

\section*{\LARGE Data availability}
The raw dataset cannot be made publicly available owing to confidentiality agreements with the facility operator. The fine-tuned model weights for MonitorVLM-v2 at the 2B and 8B scales are publicly available at \url{https://github.com/JiangWu0826/ms-swift.git}.

\section*{\LARGE Code availability}
The source code for the proposed SymPO algorithm is available at \url{https://github.com/JiangWu0826/ms-swift.git}.
% The source testing dataset is available at \url{https://github.com/JiangWu0826/ms-swift.git}.

\noindent{\bfseries \LARGE References}\setlength{\parskip}{12pt}%

\bibliography{ref.bib}
%Articles are restricted to 50 references, Letters to 30. No compound references -- only one source per reference.
%When cited in the text, reference numbers are superscript, not in brackets unless they are likely to be confused with a superscript number.
%http://www.nature.com/nature/authors/gta/#a5.4

%% Here is the endmatter stuff: Supplementary Info, etc.
%% Use \item's to separate, default label is "Acknowledgements"

\section*{\LARGE Author contributions}
J.D. and L.Z. conceived the work and supervised the project. J.W. and S.W. contributed to the theoretical derivation and the design of the framework. J.W., Y.M.and S.W. conducted the experiments and analysed the results. All the authors contributed to the writing of the paper and reviewed the manuscript.
\section*{Competing interests}
The authors declare that they have no competing interests.

\beginedfigures
% \input{suppmat.tex}

%%%%%%%%%%%%%%%%%%%%%%%%%%%%%%%%%%%%%%%%%%%%%%%%%%%%%%%%%%%%%%%%%%

\end{document}